\pdfoutput=1
\documentclass[11pt]{article}
\usepackage[english]{babel}
\usepackage[utf8]{inputenc}
\usepackage[T1]{fontenc}
\usepackage[margin=1in]{geometry}

\usepackage[font=scriptsize,labelfont=bf]{caption}
\usepackage{booktabs}
\usepackage{amsmath}

\usepackage{graphicx}
\graphicspath{{images/}}
\usepackage[export]{adjustbox}[2011/08/13]

\usepackage{hyperref}
\usepackage{color}

\usepackage[style=apa, maxcitenames=3]{biblatex}
\usepackage{csquotes}
\newcommand{\Lmanuscript}{\mathcal{L}}

\title{\vspace{-1.5cm}Using Knowledge Distillation to improve \\ interpretable models in a retail banking context}

\author{Maxime Biehler, Mohamed Guermazi and Célim Starck\\
\texttt{BPCE SA, Regulatory AI Department}\\
\small{\texttt{\{maxime.biehler, mohamed.guermazi, celim.starck\}@bpce.fr}}}

\date{}

\begin{document}

\maketitle

\begin{abstract}
    \textit{This article sets forth a review of knowledge distillation techniques with a focus on their applicability to retail banking contexts. Predictive machine learning algorithms used in banking environments, especially in risk and control functions, are generally subject to regulatory and technical constraints limiting their complexity. Knowledge distillation gives the opportunity to improve the performances of simple models without burdening their application, using the results of other - generally more complex and better-performing - models. Parsing recent advances in this field, we highlight three main approaches: Soft Targets, Sample Selection and Data Augmentation. We assess the relevance of a subset of such techniques by applying them to open source datasets, before putting them to the test on the use cases of BPCE, a major French institution in the retail banking sector. As such, we demonstrate the potential of knowledge distillation to improve the performance of these models without altering their form and simplicity.}
\end{abstract}

\vspace{.25cm}


\section{Introduction}

\subsection{Predictive models traditionally used in retail banking}
\label{retail_banking_models_constraints}
Although the banking sector holds massive troves of data regarding its customers, products and transactions, and is no stranger to using quantitative tools to inform its decisions, two constraints usually weigh on the development of predictive models.

The first one lies in the regulatory obligation to use interpretable models for a wide range of issues, with the management function being able to explain both the way a model was trained and why specific decisions have been made. Indeed, the \textcite{EBA_GL_loan} urges banking institutions to "\textit{understand the models used, and their methodology, input data, assumptions, limitations and outputs}".

The second has to do with the production environments available to deploy the models on. Due to the persistence of legacy systems, cost constraints or execution time limits --- think real time e-commerce fraud detection --- models may be limited to simple operations and conditions, i.e. a set of rules rather than a random forest, light computations in place of a fully fledged neural network.

Modeling for retail banking use cases means dealing with both these strong customers protections --- enforced through regular audits --- and the high data volume which at times shortens the time allocated to each sample. These shackles help explain why modeling practices in retail banking departments are centered around simple and interpretable models such as the logistic regression or the (shallow) decision tree.

\subsection{The trade-off between accuracy and transparency}
Not only do the models need to be transparent --- i.e. simple and interpretable --- they should also prove precise. Accuracy and transparency are usually introduced as antagonistic goals, with the implicit assumption that larger, fancier models should be able to accomplish at least as much as their simpler counterparts, if not much more. This, of course, comes at the price of the immediate interpretability of the models' outputs. It should be noted that this traditional dichotomy between the accurate black box model and the not-so-accurate transparent one does not always hold (\cite{Rudin2019Why}); in either case, common sense would dictate that simpler alternatives should always be put to the test to highlight the trade-off they make (or lack thereof). The relevance of knowledge distillation methods is a somewhat independent issue as teacher models need not always be overly complex black boxes. Indeed, gains sometimes can be secured with self distillation only (\cite{yuan2019revisit}).

\subsection{Knowledge distillation as a way to improve simple models} \label{simple_models}
The idea behind knowledge distillation is the transfer of \textit{dark knowledge} from a teacher model to another (student) model. This implicit information, resulting from the training process of the teacher model on the available data, can be used as learning material by the student model, in hopes of improving its performance. Once the latter is trained, the former is no longer needed, which limits the computing power and memory required at execution time.
In our context, our work will focus on improving the performance of the decision tree, which is generally agreed to be an interpretable model (\cite{molnar2019interpretable}). Teacher models, on the other hand, need only comply with a few operational constraints (see section \ref{3_soft_targets}). What is especially interesting about knowledge distillation in this context is that it is mostly model-agnostic and  adapts to our settings, as the student trees are able to operate on their own after their initial training.

\subsection{Plan}
The second section synthesizes the existing literature on knowledge distillation with an emphasis on the vanilla applications of regression and classification on structured data, which are the primary focus of predictive models in banking contexts. The third elaborates on the empirical setup we have used to assess the performance of different algorithms, while the fourth reports on the results we have obtained. Section five focuses on the applicability of these techniques to typical banking models with illustrations and results obtained from in-house datasets. The sixth and final section concludes on the results we have obtained and further leads to be followed.

\section{Related Work}
The idea of knowledge distillation is quite recent, with the first reference to the term being made by \textcite{hinton2015distilling}, but links to other domains exist both in terms of objectives and methods.

\subsection{Multiple flavors of knowledge distillation}
Dark knowledge refers to information not directly encoded in the original training dataset, which nevertheless is relevant to the prediction task at hand. It is made explicit by a teacher model, then passed down through knowledge distillation. Although the term has been coined in the frame of \textit{soft targets} (\cite{hinton2015distilling}), we argue that the teacher-student framework, along with the general idea of transmitting information from the one to the other through model outputs, can in practice be used in other settings while still being referred to as \textit{knowledge distillation}. While the available body of research for those is thinner, other approaches are thus included in this paper, and will be detailed alongside the seminal technique.

\subsubsection{Soft targets}
We group under the generic designation \textit{soft targets} the range of techniques starting with and expanding on \textcite{hinton2015distilling}, based on the use of a teacher model's probabilistic outputs instead of hard labels to train a simpler student model for a classification task.

The original idea is as follows: after the teacher model --- an ensemble of neural networks or a large neural network --- has been trained, it is applied back to the training dataset, which results in a matrix of predicted probabilities for the belonging of each observation to each class, which are subsequently called soft targets. 

With a softmax activation function, these soft targets can be construed as a probability distribution; for each observation, these probabilities $p^{T}_{i}$ --- where $i$ denotes the class --- are calculated as follows: 
\begin{equation} \label{soft_target_expr} 
    p^{T}_{i} = \frac{exp(z_{i})}{\sum_{j}exp(z_{j})}
\end{equation} 
with $z_{i}$ the \textit{logits} --- the values resulting from the path of the observation through the neural network and arriving at the position of the $i^{th}$ neuron in the output layer before the application of the softmax activation function. 

Two possibilities are then outlined for the training of the student model.
\label{kd_targets_loss}

In the first variant, the student model is trained with the same training dataset observations, with the difference that the target variable, initially consisting of hard targets in $\{0,1\}^{K}$ (where $K$ is the number of classes), has been replaced with these soft targets belonging in $[0,1]^{K}$.

In the second one, the student --- which is also assumed to be a neural net in the aforementioned paper --- is trained using the exact same observations, but a different loss $\Lmanuscript$, which takes into account both the original hard targets $y$ and the new soft targets $p^{T}$:

\begin{equation} \label{loss}
    \Lmanuscript=\alpha H(p^{T},p^{S}) + (1-\alpha)H(y,p^{S})
\end{equation}
with

\begin{equation} \label{cross_entropy}
H(p,q) = \sum_i{-p_i}{log(q_i)}
\end{equation}

where $H$ is the cross-entropy function, $p^{S}$ is the student model's output and $\alpha$ is a weighting parameter, which \textcite{hinton2015distilling} observe is best set close to 1.

In the wake of this publication, a number of variations of the original method have been put forward to further improve the performance of student models. Following \textcite{ba2014do} with the matching logits idea, \textcite{liu2018improving} implement a slightly different version of soft targets by using logits $z_{i}$ instead of probabilities $p^{T}_{i}$. Another variation consists in combining soft and hard targets (\cite{bai2019rectified}) with the purpose of finding an optimal intermediary configuration before training the student on it.

The work of \textcite{improve_KD} introduced two other variants to the soft targets methods family. The first one, \textit{Probability Shift} (PS), aims to swap the probabilities of the ground truth and the predicted class for misclassified samples, thus ensuring the student model never learns on its teacher's mistakes. The second one, \textit{Dynamic Temperature Distillation} (DTD), introduces a temperature adjustment (see section \ref{label_smooth}) at the observation level, depending on the student's ability to correctly handle it; the output softmax temperature is increased for confident predictions and decreased for uncertain ones. This uncommon consideration for the student's outputs is also found in \textcite{dhurandhar2018leveraging}, which we acknowledge in section \ref{sample_selection_lit}.\\

\label{label_smoothing_mention}
Why does knowledge distillation using soft targets work? The available body of research hints at a few possible reasons. \textcite{tang2020understanding} argue that the benefits of the technique are threefold: it provides label smoothing regularization --- preventing overfitting --- reweights observations to favor the most representative of each class, and supplies the student with the relationship between classes in the case of multiclass classification.

The benefits of regularization are sometimes the focus of the work, witness self-distillation (\cite{mobahi2020selfdistillation}) and early-stopped knowledge distillation (\cite{Cho_2019_ICCV}). These findings hint at an easily observable corollary: if the teacher's performance and confidence in its outputs are too high, soft targets begin to look like hard ones, and the point of using knowledge distillation is lost.

\subsubsection{Sample selection}
\label{sample_selection_lit}
A second class of methods can be gathered under the term \textit{sample selection}. The aim here is to select or weight observations in the training dataset according to the results obtained by a teacher model, so as to optimize the learning of a student model. Interestingly, \textcite{tang2020understanding} argue that such a reweighting already happens implicitly when using the \textit{soft targets} technique. It can also, however, be made explicit by other methods, which are the focus of this section.

\textcite{dhurandhar2018improving} devised such a scheme with the \textit{ProfWeight} approach. Suited to classification tasks, it consists in measuring the prediction difficulty of each observation in the training dataset through probes attached to the intermediate layers of a teacher neural network. This difficulty is in turn converted into a weight --- higher for easier samples and lower for harder ones --- which are then provided to the student model during its training phase. This effectively prioritizes the correct prediction of the observations most representative of their class so as not to overwhelm the simpler model's generalization ability. In a follow-up work, \textcite{dhurandhar2018leveraging} describe \textit{SRatio}, a formalization of their initial method which relaxes the neural net teacher constraint by introducing the concept of graded classifiers to mimic the former's intermediate layers. \textit{SRatio} also takes into account the student's predicted probability for the true class in order to overweight predictions which are deemed easy by the teacher model and hard by the student one.

Contrasting with this selection of easy and representative observations, \textcite{ghose2019learning} put forward another scheme in which the student model is exposed to a mix of easy predictions to learn the stereotypical examples of each class and hard ones to learn about their boundaries. In addition, observations are sampled instead of weighted, and their classification difficulty is gauged by the simple prediction error of a teacher model.

\subsubsection{Data augmentation}
Last but not least, another way to transfer knowledge between teacher and student is to take advantage of the former's generalization ability to make the latter explore a larger portion of the observations' feature space than is initially afforded by the training data.

Manufacturing synthetic training data for this purpose was promoted as early as the nineties when \textcite{breiman1996born} outlined the \textit{born again trees} method, which aims at building a single better tree predictor by using the outputs of a tree ensemble. After the usual training of the latter, their idea leans on a \textit{smearing} technique designed to produce similar yet slightly different samples than those existing in the training set. These samples are labeled with the help of the ensemble, then a single tree is grown and pruned using only these artificial observations. In a similar fashion, \textcite{bucilua2006model} reported performance improvements for a single neural network with their own framework to create and label synthetic data points using an ensemble of nets.

Instead of manufacturing data points, one could also make use of the heaps of unlabeled data that sometimes exist alongside the usual richer training set. \textcite{mosafi2019deepmimic} demonstrated this point by having a student neural net learn from the outputs of a teacher on a simulation of unlabeled data consisting of training observations stripped of their label. Interestingly, their scheme also makes use of soft targets in that the student model is trained on probability distributions.

\subsection{Links to other research domains}
Many fields of research are interwoven with knowledge distillation, which helps to explain how it came to be, informs its future developments and provides new application possibilities.

\paragraph{Model Compression}
is a field dedicated to reducing the size of models, with the typical application of enabling the deployment of complex and accurate models (such as neural networks) in low-performance devices (such as smartphones). Knowledge distillation is deeply rooted in model compression --- witness the works of \textcite{breiman1996born}, \textcite{bucilua2006model} or \textcite{hinton2015distilling}, all concerned with the replacement of an ensemble with a single model --- and sometimes even considered a subfield of the latter (\cite{cheng2018recent}, \cite{cheng2018model}). \textcite{ba2014do} offer yet another example of this proximity, by training shallow neural nets to \textit{mimic} an ensemble of deep convolutional models through the use of \textit{matching logits}. 

While the two fields are closely related, they exhibit subtle differences. Model compression, on the one hand, appears to be mainly concerned with staying within the same model type or family (e.g. pruning or quantization techniques for neural networks) and minimizing the loss of performance compared to the complex model. Knowledge distillation, on the other hand, allows to cross the borders of model families more easily, and focuses on the performance improvements for the weaker model.

\paragraph{Model Interpretability}
is a requirement that has become more and more common as machine learning is being deployed in sensitive domains such as healthcare (\cite{miotto2017deep}) or justice (\cite{wexler2017computer}), where understanding how a model functions and why it came to its conclusions is just as important as the results themselves. While there is no universal definition of model interpretability (\cite{doshivelez2017towards}), complexity buildup (e.g. deep neural nets) or trade secrets (e.g. proprietary models) are known ways to impair it. A key distinction in this field separates intrinsic interpretability --- choosing and training easy-to-understand models --- and post hoc interpretability --- explaining their predictions after the fact (\cite{molnar2019interpretable}). Knowledge distillation somewhat blends both sides by providing a standalone, intrinsically interpretable model that may also be used to provide explanations for the teacher model's outputs. Indeed, it allows for the training of a simple model which strives to mimic the output of a more complex one over the complete feature space and fits nicely in the interpretability toolbox as a way to develop \textit{global surrogates} of complex models. However, knowledge distillation may not be enough for interpretation purposes, as the training of a complex model and its student creates, in itself, a convoluted pipeline. Furthermore, the inability of the student model to perfectly reproduce the teacher's outputs casts doubt on the explanation and the original prediction, as argued by \textcite{rudin2019stop}.

The issue of model interpretability is not new but made a recent resurgence due to the progress made with the notably opaque neural networks and their expansion into new application domains. It is becoming all the more important as recent regulations developments in Europe grant citizens the right to an explanation when their data are subject to automatic processing (\cite{EU2016GDPR}) and, in certain contexts, when decisions affecting them are made by an algorithm (\cite{EU2020AIEthics} --- not yet binding).\\

Most of the published research is concerned with state of the art (read complex) models and tasks, such as picture or text recognition, for which even the student model can be quite complex (e.g. a thinner, shallower neural network). Our endeavor is, in comparison, quite simpler, as we limit ourselves to interpretable student models --- see section \ref{simple_models}.

Most of the existing works also focus on classification tasks, but adjustments to regression problems are possible outside of the \textit{Soft Targets} realm, and sometimes straightforward, as we illustrate in section \ref{data_aug}. The \textit{Data Augmentation} technique can be applied to the the regression cases, as well as the sample selection for which we can use the error as metric to distinguish the easy-to-predict observations from the others. A few research papers --- such as \textcite{takamoto2020efficient} --- have looked into the application of knowledge distillation to regression models.

\section{Empirical Setup} \label{empirical_setup}
Based on the constraints mentioned in section \ref{retail_banking_models_constraints}, we focus on a selection of knowledge distillation techniques to implement. We also provide a description of the open data sets used to evaluate these algorithms before applying them to retail banking use cases.

All tests were conducted with Python, using the Keras library (\cite{chollet2015keras}) for neural networks and the scikit-learn library (\cite{scikitlearn}) for other models.

\subsection{Datasets}
\label{datasets_description}
This section introduces the open data sets we used as benchmarks to evaluate our models.

\begin{table}[!htbp]
\caption{Datasets dimensions.}
\label{tab:datasets}
\begin{center}
\begin{tabular}{lccc}
\toprule
\multicolumn{1}{l}{Dataset} & Class & Features & Instances \\ \midrule
Adult                       & 2     & 14       & 48,842     \\
Connect-4                   & 3     & 42       & 67,557     \\
MNIST                       & 10    & 784      & 70,000     \\
SGEMM                       & \textit{reg}   & 14       & 241,600   \\ \bottomrule
\end{tabular}
\end{center}
\end{table}

\paragraph{Adult} contains observations for 48,842 employees, including 14 attributes such as age, marital status, education, working hours, etc., and a target boolean variable indicating whether their annual compensation exceeds \$50,000 or not.  As such, it lends itself to a binary classification problem (\cite{Dua:2019}).

\paragraph{Connect-4} is a dataset in which each of the 67,557 observations represents the 6-by-7 grid occupation in the eponymous game, along with the result of the duel --- a three-class variable that includes the possibility of a draw. The features thus correspond to the 42 locations in the grid and indicate whether each of them is occupied by Player 1, Player 2 or has remained empty (on which we apply a one-hot encoding before training). The target is the outcome of the game (\cite{Dua:2019}).

\paragraph{MNIST} consists of an ensemble of 28-by-28 grayscale pictures representing digits ranging from 0 to 9, the underlying objective being to identify the digit from the pixels data. The dataset contains 70,000 observations and 784 features corresponding to pixels (\cite{lecun-mnisthandwrittendigit-2010}).

\paragraph{SGEMM GPU kernel performance} contains measures of the time required to run a matrix product computation algorithm, along with the calculation parameters used. For each particular setting, the algorithm is launched four times so the dataset contains four timings per observation, which we average and scale down to $[0,1]$ to approach our real use case. The dataset contains 241,600 observations with 14 explanatory variables (\cite{ballesterripoll2017sobol}).

\subsection{Soft Targets}
\label{3_soft_targets}
We test all \textit{soft targets} variants with a neural network as teacher --- specifically, a Multilayer Perceptron (MLP) --- and a CART decision tree as student. After training the teacher model, we obtain soft targets as a result of applying a softmax activation function at the output layer.

Thereafter, these soft targets replace the initial target variable (e.g. the binary label for the \textit{Adult} dataset) along the original, unchanged explanatory features.

It is worth emphasizing that although many works --- including this one --- use a neural network teacher, this is not a requirement for knowledge distillation to work. As long as the teacher model outputs a probability distribution for classification tasks, and the student model can be trained with probabilities instead of hard labeled classes, the \textit{soft targets} technique is applicable.

Besides, in order to retain some flexibility in selecting the desired machine learning models, we adopt the probabilities approach instead of the losses approach (as outlined in section \ref{kd_targets_loss}) which is more suitable when both the teacher and the student models are neural networks. Given that a probability is assigned to each class, nonbinary classification problems become multi-output regression ones with a purpose of classification for students. The evaluation of the models for classification tasks then simply requires to consider the class associated with the maximum probability:

\begin{equation} \label{argmax}
y_{prediction} = argmax(y_{probabilities})
\end{equation}

\subsubsection{Label Smoothing}
\label{label_smooth}
As mentioned in section \ref{label_smoothing_mention}, \textit{Label Smoothing} is usually used to improve the performance of a model by preventing it from overfitting the training data, and the soft targets framework already include such a correction. Nevertheless, one can try to go further in this direction and apply additional smoothing to the targets before passing them on to the student model.

As our teacher model is a neural network, the usual way of implementing label smoothing consists in increasing the temperature $T$ of the softmax activation function in its output layer. Temperature is an optional parameter in the softmax activation function, which divides the logits:

\begin{equation} \label{temperature}
    q_{i} = \frac{exp(\frac{z_{i}}{T})}{\sum_{j}exp(\frac{z_{j}}{T})}
\end{equation}

\begin{figure}[tbh!]
\centering
\includegraphics[width=0.75\columnwidth]{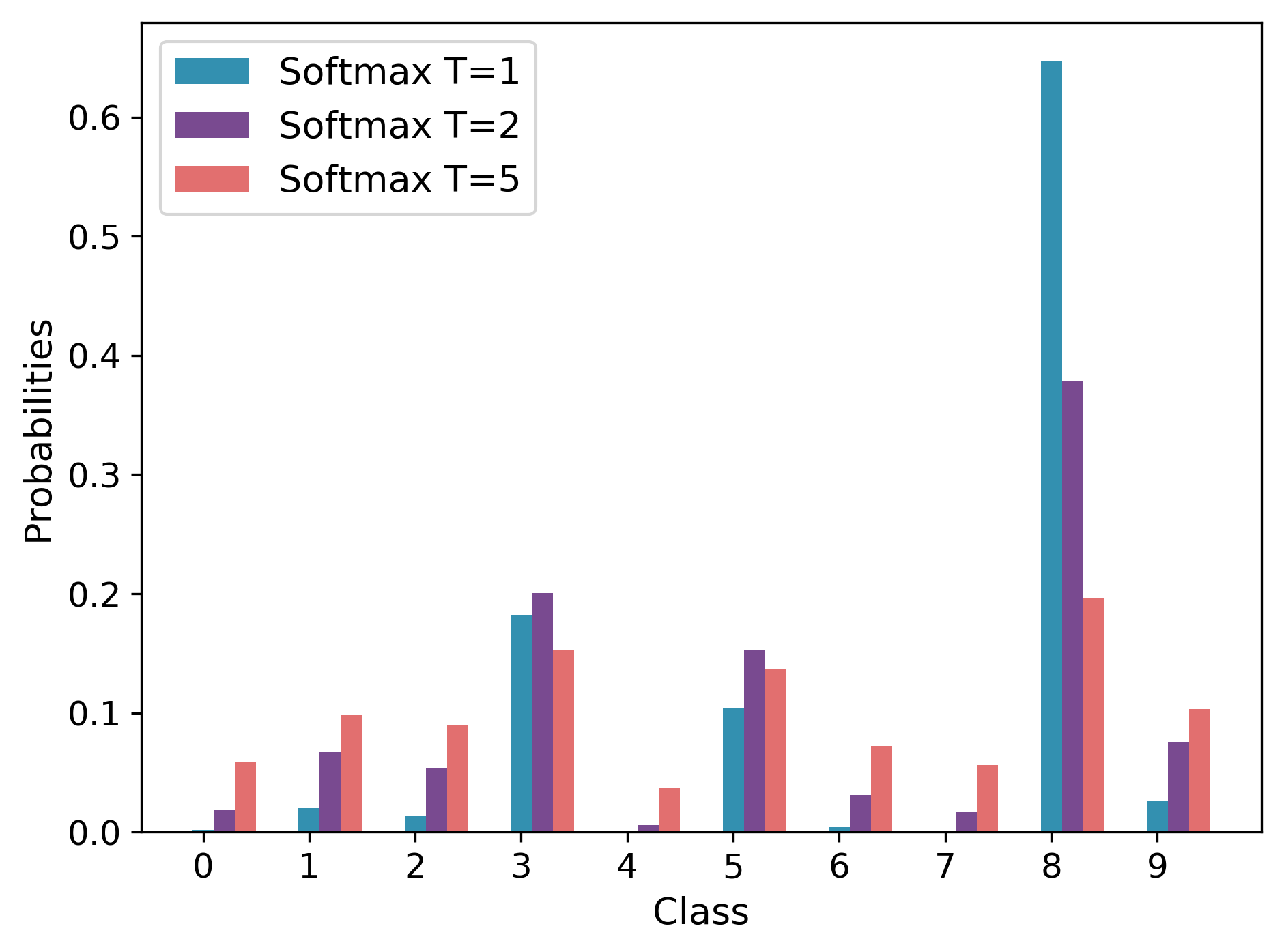}
\caption{Probability distribution of softmax function, with T=1, T=2 and T=5.}
\label{label_smoothing_temperature}
\end{figure}

In a standard application of the softmax function, the temperature $T$ is equal to 1. For smoothing purposes, $T$ can be increased in order to obtain a somewhat flatter probability distribution that helps avoiding overconfidence and overfitting. As $q_i$ tends towards a uniform probability distribution, $\frac{1}{K}$ (where K is the number of classes) when $T$ approaches infinity (Figure \ref{label_smoothing_temperature}), the extent of temperature increases should be limited lest information is lost. In our experiments, we set $T=5$.

\begin{equation} \label{softmax_infty}
    q_i \underset{T \to +\infty}{\overset{}{\longrightarrow}} \frac{1}{K}
\end{equation}

Interestingly, \textcite{label_smoothing_help} witnessed that in a context of knowledge distillation, label smoothing applied to the teacher can hurt the student's performance. In our work --- with teachers trained using hard targets and their outputs softened through temperature or linear adjustment --- we haven't noticed any such degradation.

\subsubsection{Mixed Labels}
One can also go the other way around and reduce the soft labels uncertainty by mixing the teacher's outputs back with the original labels. The idea behind this method is to reduce the loss of information, and possibly correct the teacher's mistakes while still using a soft enough target. \textcite{bai2019rectified} put forward a simple linear combination to implement this idea:

\begin{equation} \label{mixed_label}
    y_{new}=\alpha\ y_{hard}+(1-\alpha)\ y_{soft}
\end{equation}

where $\alpha$ is a real hyperparameter in $[0, 1]$ which we adjust for each model using a validation sample.

\subsubsection{Probability Shift}
\begin{figure}[tbh!]
\centering
\includegraphics[width=0.75\columnwidth]{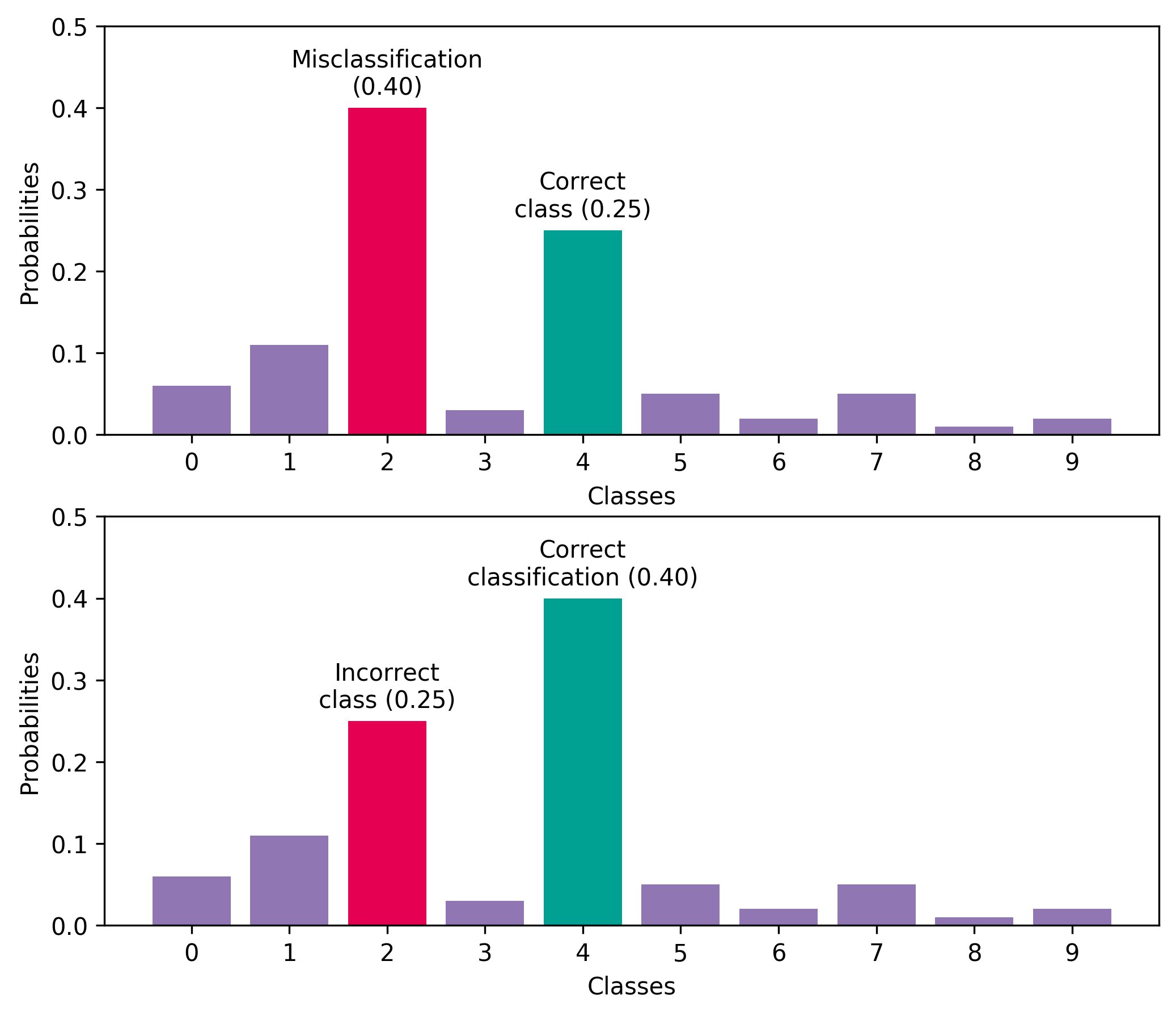}
\caption{Teacher prediction probability histogram for a misclassified MNIST sample before (top) and after (bottom) Probability Shift.}
\label{chart_probabilityshift}
\end{figure}

As the other soft targets variants, \textit{Probability Shift} is straightforward to implement. Once the teacher has been trained and applied back on the training data, we swap the probabilities of the ground truth class and the predicted one for misclassified samples, as illustrated on the MNIST dataset by Figure \ref{chart_probabilityshift}. The student is then provided with the corrected matrix of targets (\cite{improve_KD}).

\subsubsection{Matching Logits}
Following the steps of \textcite{liu2018improving}, we also consider using logits instead of the probabilities resulting from the teacher's softmax layer, as an alternative representation of the dark knowledge being transferred.

The logits in a neural network are computed with the following formula:

\begin{equation} \label{logits}
    z = \Theta_{l}*{x'_{l-1}}
\end{equation}

where $\Theta_{l}$ is the weight matrix of the last hidden layer, and $x'_{l-1}$ the output activation matrix of the penultimate hidden layer supplemented by the bias vector.

The student model is trained on this target variable (\ref{logits}). For later prediction purposes, the softmax activation function (\ref{soft_target_expr}) is applied on the student's outputs.

\subsection{Sample Selection}
Another way to go about knowledge distillation is to keep the target variable as-is and to make the teacher model define the training set instead --- a range of methods we label \textit{sample selection}. 

Focusing on the \textit{ProfWeight} technique, we choose to use the same teachers as for the soft targets methods, namely Multilayer Perceptrons. Once an MLP has been trained on the dataset, we follow the steps outlined by \textcite{dhurandhar2018improving} to train single-layer perceptrons on the intermediary data representation of each hidden layer and derive sample weights based on the performance of these auxiliary models (Figure \ref{sample_selection_schema}).

\subsection{Data augmentation} \label{data_aug}
For the \textit{Data Augmentation} technique, we turn to a regression task using the SGEMM GPU kernel performance dataset. To better approximate our real use case (see section \ref{LGD_section}), we average its timings and scale these values with scikit-learn's min-max scaler:

\begin{equation} \label{minmax}
    Y_{new} = \frac{Y_{init}\ -\ min(Y)}{max(Y)\ -\ min(Y)}
\end{equation}

Since our dataset is fully labeled, we can either remove y-values from a portion of it, or create a new unlabeled dataset using dedicated techniques such as \textit{smearing} (\cite{breiman1996born}) or \textit{SMOTE} (\cite{Chawla_2002}). 

For the sake of simplicity, and given the comparatively large size of the dataset, we choose the former and consider that only 2\%of the samples, randomly selected, are labelled. Among these 2\%, a fifth is set aside for the evaluation part of our study while the remaining four fifths are used for training purposes ($D$). 98\% of the data is thus stripped of its target value and will play the part of an unlabeled dataset ($D'$).

We take as teacher model a MLP with 2 hidden layers, which we train on the labeled dataset $D$ and apply on the simulated unlabeled one, $D'$, adopting the predictions as new ground truth for it.

Once we are in possession of both the training dataset $D$ and the recently labeled dataset $D'$, we need to estimate the optimal amount of data we should select from $D'$ and add to $D$ (forming $D''$) before training the student. As we intend to evaluate the performances of this technique with CART depths ranging from 4 to 10, we consider a CART model of depth equal to 7 as a preliminary student model, so as to decide which portion of $D'$ should incorporate $D"$.

We evaluate this student's MSE by adding various percentages of random data from the recently labeled dataset and obtain the results in Table \ref{tab:data_augmentation}. 

\newpage

\begin{table}[!htbp]
\caption{MSE of the CART student as a function of to the percentage of randomly added data. The original training set $D$ contains 3,866 observations and 1\% of the recently labeled dataset $D"$ corresponds to 2,368 observations.}
\label{tab:data_augmentation}
\vskip 0.02in
\begin{center}
\begin{tabular}{cc}
\toprule
\% added data & Resulting MSE   \\ \midrule
0            & 24.73 \\
2            & 13.66 \\
4            & 10.31 \\
\textbf{5}   & \textbf{10.26} \\
6            & 10.48 \\
8            & 10.67 \\
10           & 11.40 \\ \bottomrule
\end{tabular}
\end{center}
\vskip -0.02in
\end{table}

We notice that as soon as we add a small percentage of the recently labeled data, the MSE decreases. From Table \ref{tab:data_augmentation}, we decide to add 5\% of $D'$ (11,838 observations) to $D$ to create the final dataset $D''$ on which the student model will be trained.\\

\begin{table*}[!ht]
\caption{Teacher and student models used to evaluate knowledge distillation methods.}
\label{tab:teacher_student_configuration}
\vskip 0.02in
\begin{center}
\begin{adjustbox}{center}
\resizebox{\textwidth}{!}{
\begin{tabular}{lcc}
\toprule
Dataset         & Teacher model     & Student model\\ \midrule
Adult           & MLP, 2 hidden layers (48, 24) & CART, at least 40 samples per leaf\\
Connect-4       & \begin{minipage}[t]{0.5\columnwidth}\centering
MLP, 3 hidden layers (256, 128, 128),\\each followed by a dropout regularization (0.8)\end{minipage} & CART, at least 40 samples per leaf\\
MNIST           & \begin{minipage}[t]{0.5\columnwidth}\centering
MLP, 3 hidden layers (256, 256, 256)\end{minipage} & CART, at least 20 samples per leaf\\
SGEMM    & 
\begin{minipage}[t]{0.5\columnwidth}\centering
MLP, 2 hidden layers (40, 20)\end{minipage} & CART, at least 40 samples per leaf \\ \bottomrule
\end{tabular}}
\end{adjustbox}
\end{center}
\vskip -0.02in
\end{table*}

All models specifications used to run the experiments are recorded in Table \ref{tab:teacher_student_configuration}.

\section{Results}
This section details the results of the aforementioned techniques, highlighting the improvements afforded by knowledge distillation. We provide the results for classification tasks first, followed by a trial of the data augmentation method for the regression use case. In the spirit of robustness, all our reported results are the average of 10 randomized runs with distinct splits of the original dataset into training and test ones.

\subsection{Classification}
In addition to the baseline CART fitted to depths ranging from 4 to 12, we assess six knowledge distillation approaches for the classification use case : standard soft targets, soft targets with label smoothing, probability shift, mixed labels, matching logits, and the profweight flavor of sample selection.
We measure the performance of each of these algorithms on the selected datasets (section \ref{datasets_description}). Full results are available in appendices.

\begin{figure}[tbh!]
\centering
\includegraphics[width=0.65\columnwidth]{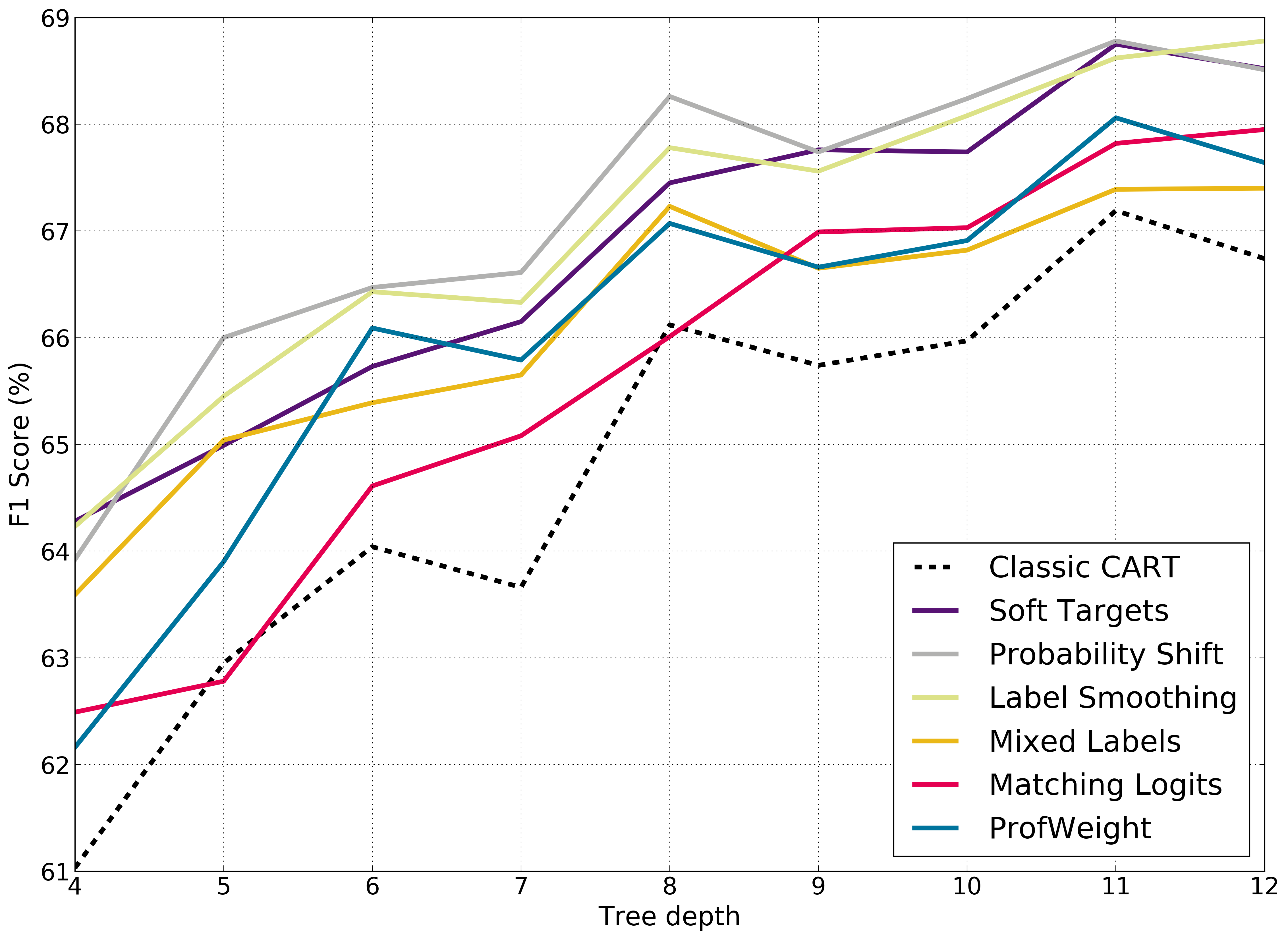}
\caption{Chart of F1 scores for \textbf{Adult} as a function of CART depth. F1 Score of teacher model (MLP): 85.5\%}
\label{adult_graph}
\end{figure}

The first dataset, \textbf{Adult}, is unbalanced. Thus, we measure the F1-score --- rather than the accuracy --- for each technique and obtain the results presented in Figure \ref{adult_graph}.

The first conclusion we can draw from this outcome is that the knowledge distillation techniques we implement generally outperform the vanilla model by 2 to 3 F1-score points. Topping the chart, the \textit{Probability Shift} variant obtains the best classification performance for 6 different CART depths out of the 9 tested, with \textit{Label Smoothing} and \textit{Vanilla Soft Targets} close seconds.

\begin{figure}[tbh!]
\centering
\includegraphics[width=0.65\columnwidth]{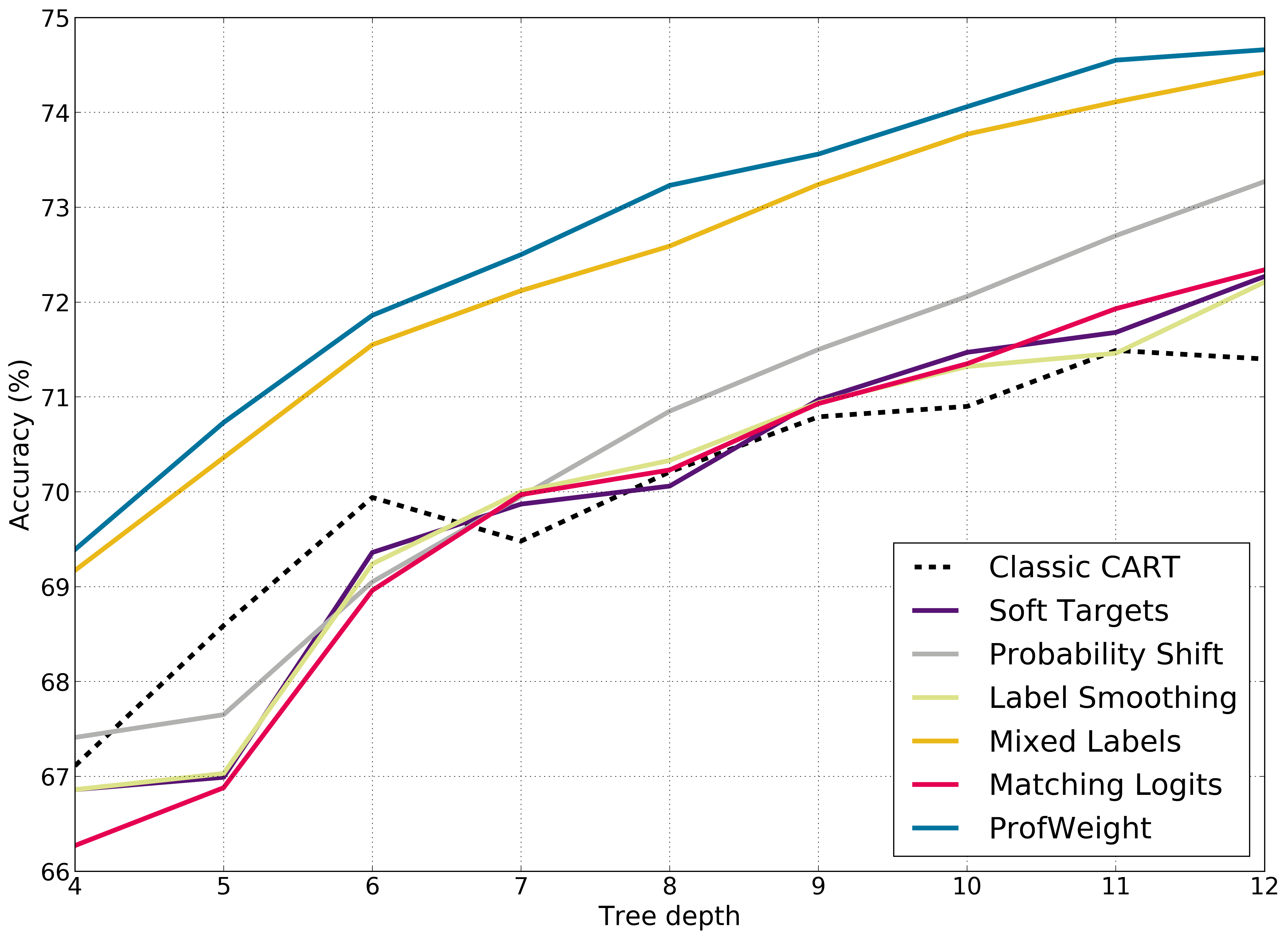}
\caption{Chart of accuracies for \textbf{Connect-4} as a function of CART depth. Accuracy of teacher model (MLP): 81.5\%}
\label{connect4_graph}
\end{figure}

We apply the same techniques on the \textbf{Connect-4} dataset, and plot our results in Figure \ref{connect4_graph}.

The chart illustrates once again that knowledge distillation techniques may offer marked improvements in performance over the vanilla student model, but the assessment should be nuanced: this time, the \textit{Mixed Labels} and \textit{ProfWeight} methods deliver a 2--3-point lead in accuracy across all depths, while other variants barely hover above the vanilla student model's accuracy for depths higher than 7.

\begin{figure}[tbh!]
\centering
\includegraphics[width=0.65\columnwidth]{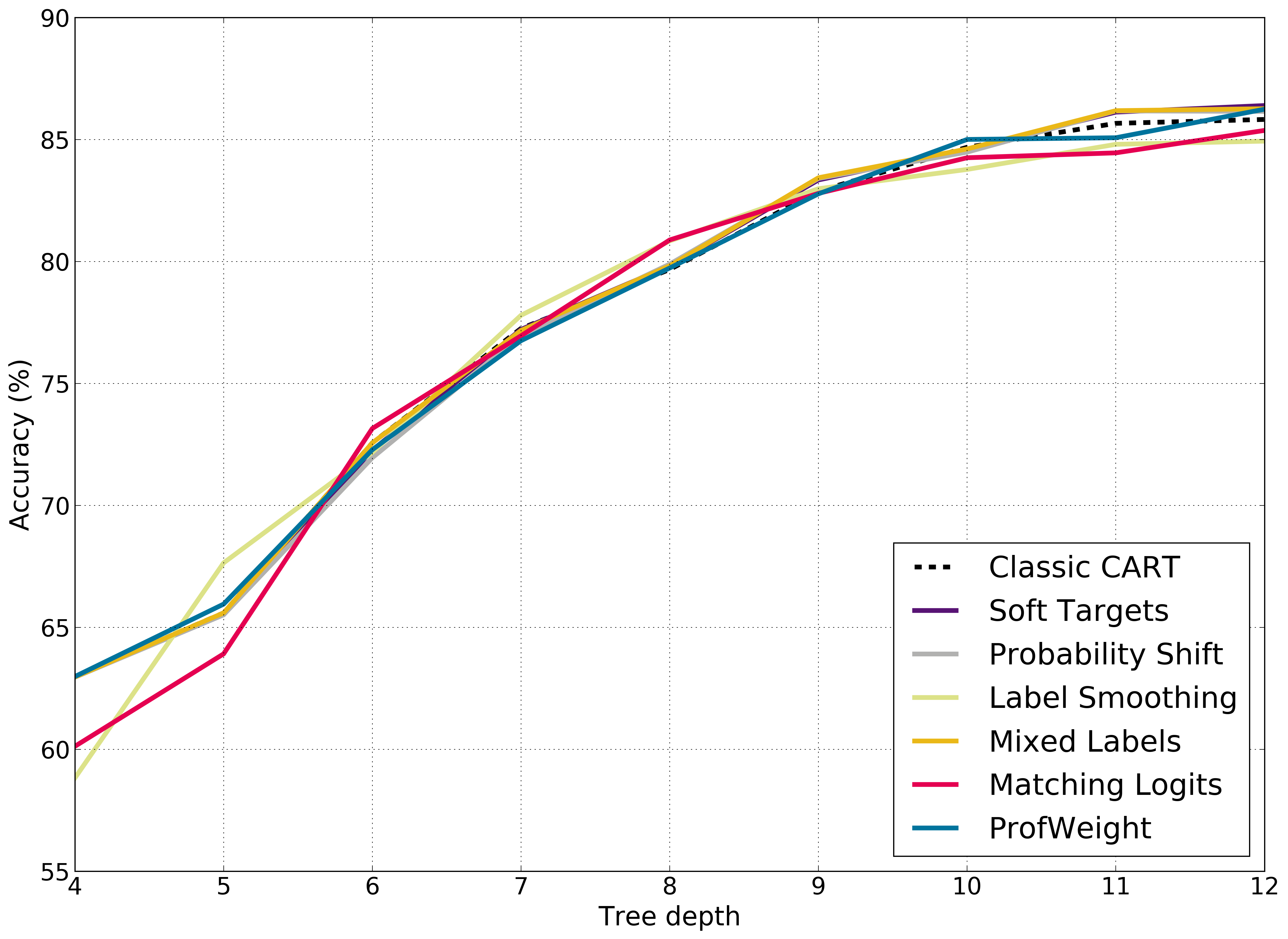}
\caption{Chart of accuracies for \textbf{MNIST} as a function of CART depth. Accuracy of teacher model (MLP): 97.13\% }
\label{mnist_graph}
\end{figure}

For our final classification task, we turn to the results obtained on the \textbf{MNIST} dataset (Figure \ref{mnist_graph}).

The chart shows that the performance gap between the knowledge distillation techniques and the vanilla student model is insubstantial here, with the \textit{Mixed Labels} model pulling slightly ahead of the pack. On the flip side, none of the tested variants perform significantly worse than the standard CART on this dataset, with hardly anything to choose between the best and worst performers --- 0.75 points of accuracy on average across all depths.\\

Our experiments show that knowledge distillation is able to boost a simple tree's performance by a few percentage points in different settings, with little to no downside. The gain is on average 2.3 points of F1-score for the Adult dataset (Probability Shift) and 2.7 points of accuracy for the Connect-4 dataset (Profweight). They also make clear, however, that there is no silver bullet as far as knowledge distillation is concerned, as no single method consistently stands out from the others. The benefits appear dependent on the use case, and especially the dataset.

\subsection{Regression}
To assess the relevance of knowledge distillation for regression purposes, we evaluate the \textit{data augmentation} technique on the \textbf{SGEMM GPU kernel performance} dataset, adding 5\% of the teacher-labeled data to an otherwise untouched training set.

\begin{figure}[tbh!]
\centering
\includegraphics[width=0.65\columnwidth]{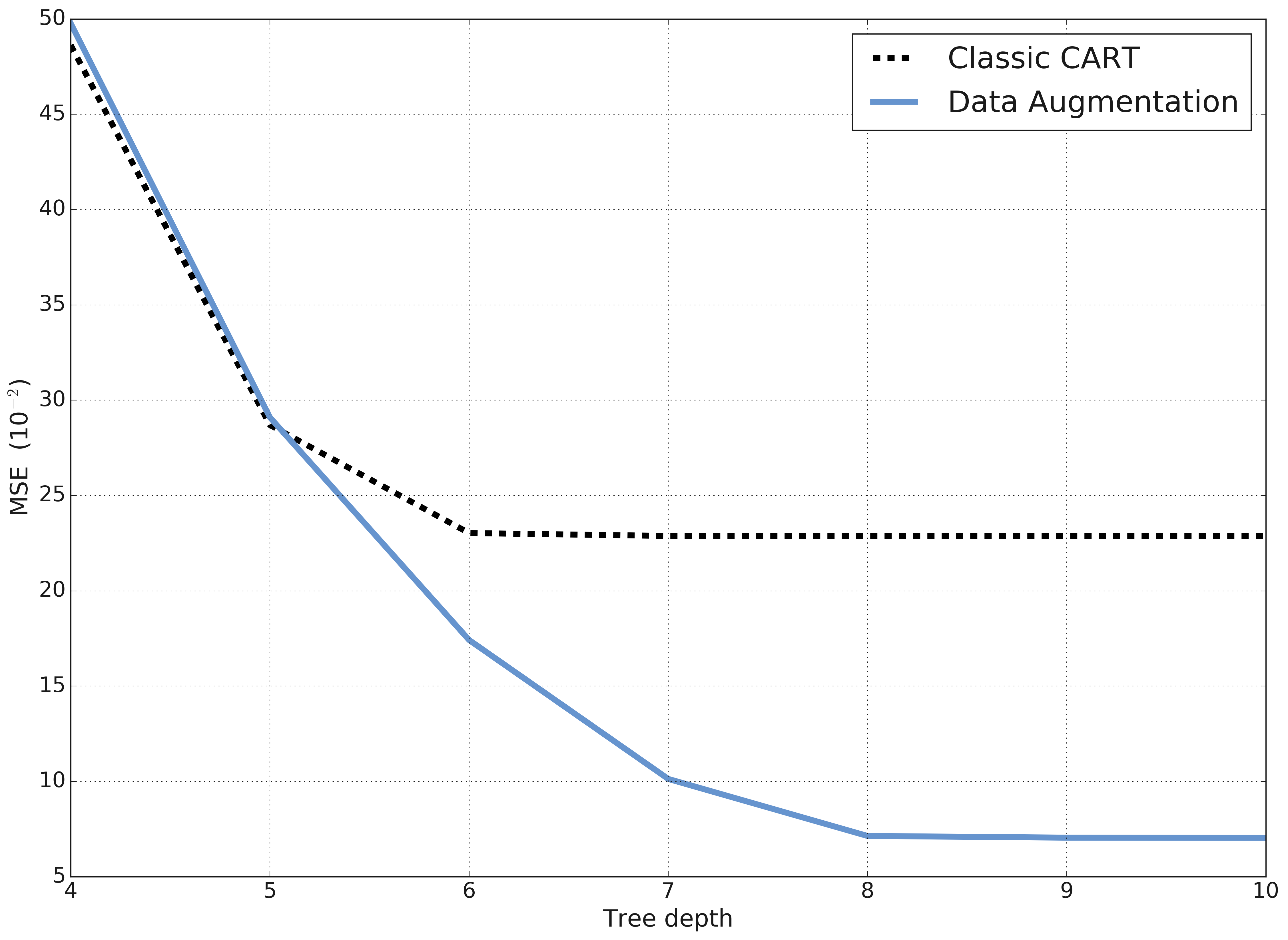}
\caption{Chart of MSE for \textbf{SGEMM} as a function of CART depth.}
\label{sgemm_graph}
\end{figure}

As shown in Figure \ref{sgemm_graph} and Table \ref{tab:SGEMM_results}, the student model MSE decreases considerably with this addition, although the difference only starts showing for depths higher than 5.

The vanilla CART model MSE decreases with the augmentation of the tree depth, up to 6, before the error levels. The student tree, on the other hand, continues to improve up to a depth of 8 before plateauing, which suggests that it benefits from the additional data.

This variant of knowledge distillation may prove useful when setting up a regression model with a training dataset of limited size. If no unlabeled dataset is available, new observations may be created with the techniques mentioned in section \ref{data_aug} and labeled using the teacher model.

\section{Application to banks' risk models}
While the positive results we obtained with open datasets are satisfactory and provide us with an overview of knowledge distillation methods, our real interest lies in the improvement of models used in retail banking. Accordingly, this section focuses on a couple of applications of knowledge distillation to banks' risk models using datasets provided by BPCE.

For the sake of fairness, our experiments are based on statistical approaches only, as we intend to test new techniques and compare them with existing ones on an equal footing. Expert model adjustments such as the restriction of variables, the simplification of thresholds, and other human interventions which are usual for these use cases, are thus not considered.

We put forward two distinct retail banking applications: credit card fraud detection in online transactions --- for which we evaluate the \textit{Soft Targets} and \textit{ProfWeight} techniques --- and Loss Given Default estimation --- for which \textit{Data Augmentation} is a natural fit given the data available for defaulted and non-defaulted contracts.

\subsection{Credit card fraud detection}
\subsubsection{Context}
Given the increasing part of online sales, credit card fraud is a growing threat with far-reaching consequences for financial institutions, faced with either taking the corresponding loss or handling a disgruntled customer.

Indeed, hacked companies and data leaks, not to mention simple thefts, provide criminals with material galore to impersonate real clients on shopping sites and pay using their credit card. During such an online sale, the customer's bank can try and detect fraudulent activity using the information at its disposal, but only has a split second to do so and accept or reject the transaction --- thus limiting the possible complexity of the underlying model tasked with this decision. As might be expected, credit card fraud detection and prevention is tremendously challenging for banks.

Our purpose in applying knowledge distillation to this problem is to improve the performance of a decision tree by detecting more fraudulent transactions while limiting false positives, which trigger unwarranted verification calls and are another source of customer inconvenience.

\subsubsection{Dataset}
Our dataset consists of hundreds of millions of transactions. Each observation records data related to the transaction as well as the binary target variable indicating whether the transaction has been reported fraudulent. The dataset is highly unbalanced as there is a dominance of non-fraudulent transactions and only about 0.1\% of fraudulent ones.

In order to reduce the amount of data to a reasonable number of homogeneous observations that we can process with a single model, we focus on a single acceptor (online shop) for which there is a fairly significant number of recorded frauds, a major French online retailer in the transportation sector. We also restrict the dataset to a short period of time (from November 2016 to March 2017) because fraudsters tend to change their techniques over time. The first three months are used as training set, one month is dedicated to validation and one month to testing. After preprocessing the data, the number of observations has dropped to just over one million, from which only 0.07\% are frauds.

\subsubsection{Tests and Results}
We train a MLP teacher on our dataset, then our student model using \textit{Soft Targets} and \textit{ProfWeight}, and compare the results with a classic CART, but also with rebalancing techniques --- which are the usual answer to imbalanced datasets. We therefore tested under-sampling (the majority class) and over-sampling (the minority class) separately, with the objective of having a fraud proportion of 0.2\% instead of 0.07\%. By choosing such a ratio, we minimize the risk of losing too much information in the first case and that of constructing too many false observations in the second case.

To evaluate the performances of these techniques, our metrics of interest are the \textit{weighted recall} and the \textit{F1 hybrid score}. This choice is motivated by the fact that it is more relevant to consider the recall in relation to the corresponding transaction amounts, so as to be representative of the money the model could prevent from being defrauded. The precision, however, denotes the customer inconvenience generated by verification calls, and as such should remain linked to the number of transactions. This leads to a hybrid F1 score which takes into consideration both the minimization of false alerts and the fraud amount correctly identified.

For the weighted recall, we take into account the amounts of the fraudulent transactions: 

\begin{equation}
\begin{gathered}
Recall_{weighted} =  \frac{amounts(TP)}{amounts(TP+FN)}
\end{gathered}
\end{equation}

with TP the true positives and FN the false negatives among the predictions of the model. Thereafter, the F1 hybrid score is given by:

\begin{equation} 
\begin{gathered}
F1_{hybrid} = \\ 2*\frac{Recall_{weighted}\ *\ Precision}{Recall_{weighted}\ +\ Precision}
\end{gathered}
\end{equation}

\begin{figure}[tbh!]
    \centering
    \includegraphics[width=0.75\columnwidth]{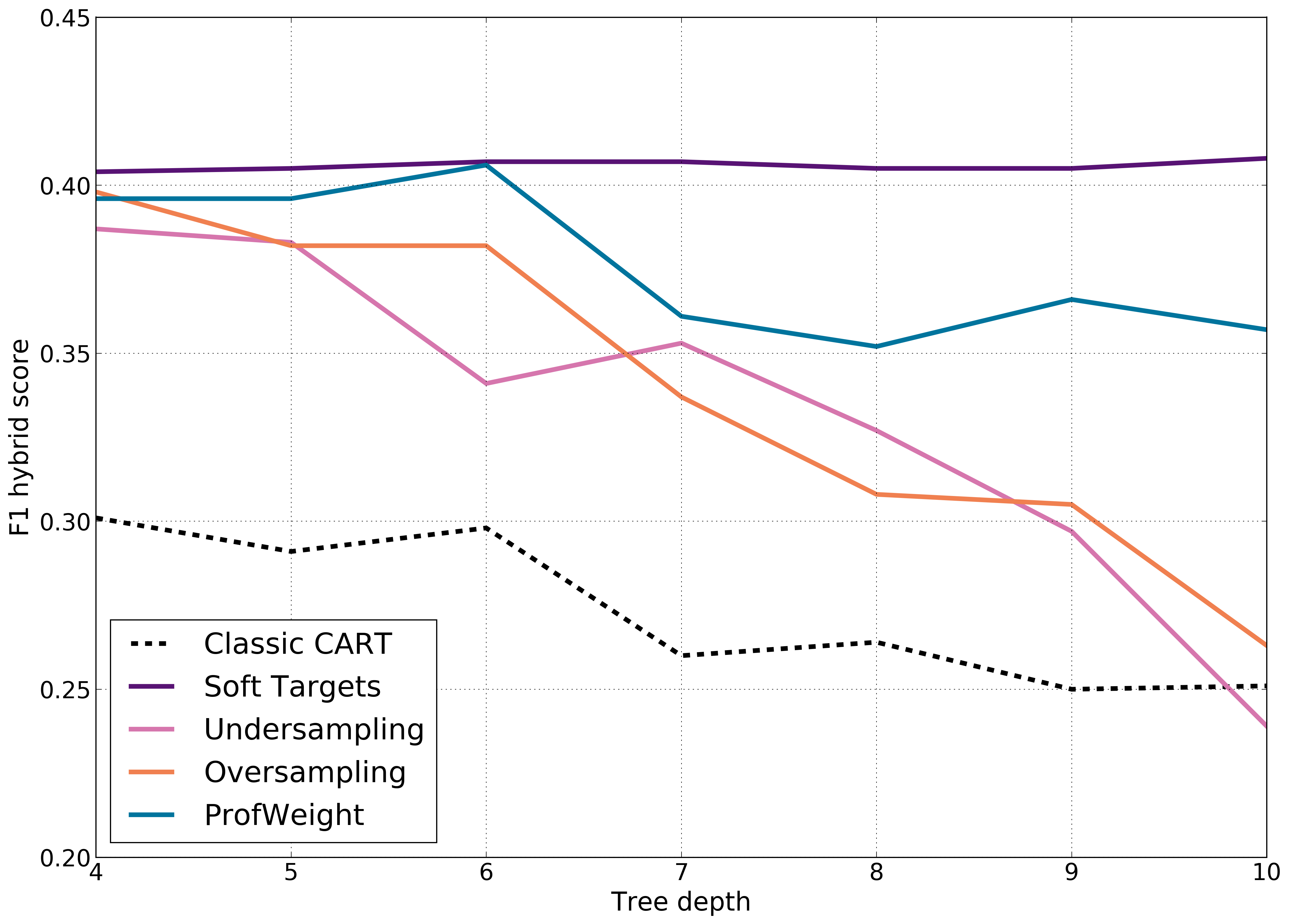}
    \caption{Chart of F1 hybrid scores for fraud prediction as a function of CART depth. Performance of teacher model (MLP): 0.412}
    \label{fraud_prediction}
\end{figure}

Figure \ref{fraud_prediction} shows a comparison of the results of the vanilla decision trees, the student ones, and the trees trained on resampled data, with full results available in appendix \ref{results_fraud}. Soft targets student models noticeably outperform the standard trees by more than 10 points of hybrid F1-score across all depths. We also note that the results of the student models come close to those of the teacher MLP model. Interestingly, resampling techniques significantly improve the decision tree performance, yet the student models still pull ahead. This is especially true at higher depths, where overfitting may play a role in the degraded results of the trees given the limited set of explanatory variables, illustrating once again the regularization benefits of knowledge distillation.

\subsection{Loss Given Default (LGD) prediction} \label{LGD_section}
\subsubsection{Context}
Credit risk is another typical use case of predictive modeling in banks. The Basel II framework (\cite{BIS2004BaselII}) introduced 3 key metrics for the purpose of measuring this risk: the Probability of Default (PD), the Loss Given Default (LGD), and the Exposure At Default (EAD).

For the computation of these parameters, mathematical models have been set up in compliance with the standards of the Basel Committee and numerous other regulatory texts which mandate, among other requirements, the use of interpretable models whose outputs can be explained by banks (\cite{EBA_GL_loan}).

In this context, we choose to focus on the estimation of LGD, with the expectation that knowledge distillation could help improve and refine predictions while keeping the model in the form of a decision tree. Besides accuracy, the homogeneity of groups of observations handled in a similar fashion is of particular interest, as the concept of \textit{homogeneity of risk classes} is underlined in the regulations relating to this type of model. It is therefore a modeling objective in its own right: the more homogeneous the leaves, the better the tree. Finally, taking into account that one of the main limiting factors of this problem is the lack of data (it is only possible to train a model on defaulted clients), we decide to apply the \textit{Data Augmentation} technique.

\subsubsection{Dataset}
Our experiment draws upon two datasets, named after the convention laid out in section \ref{data_aug}. The first one, $D$, holds the data of defaulted contracts, including the characteristics of both the customer and the product, and our target variable, the observed LGD --- which makes this use case a regression task. The second one, $D'$, contains the same information for solvent customers' contracts, without any observed LGD data. Healthy banks \textit{usually} have more solvent customers than defaulted ones, which means that $D'$ is much larger than $D$.

Since $D$ and $D'$ represent a large amount of data encompassing several types of banking products, we conduct our analysis on one product type: \textit{Personal Accounts}.

Furthermore, in order to limit censorship of the recovery process and to observe \textit{final} LGDs only, we have also restricted our labeled dataset to recovery processes observed long enough to exceed the maximum recovery horizon of the product (54 months) or those declared closed.

These steps result in a first dataset of 890 thousand defaulted contracts and their observed LGD, and a second one of 5.8 million unlabeled observations ($D'$). Both datasets share the same 72 features.

\subsubsection{Tests and Results}
First, we set up an MLP as teacher model with three hidden layers and a Rectified Linear Unit (ReLU) activation function. Consistent with common practice, we choose the CART decision tree as student model.

A quarter of the labeled dataset is retained as test dataset and the rest serves to train the neural network ($D$). The test dataset will be used to compare the performances of the classic  and student decision trees.

Thereafter, we apply the teacher MLP to the unlabeled $D'$ dataset. The results we obtain are added back to the dataset as the new targets for these solvent clients. We then take the original labeled dataset $D$ and add the equivalent of 25\% of its size in newly labeled $D'$ data, resulting in a larger $D''$ dataset.

Finally, we train student decision trees on $D''$ for depths ranging from 4 to 12. To compare the results, we examine the distribution of prediction errors, and the proportion of deviations below 0.05, 0.1 and 0.2.

Another criterion to assess whether our student decision trees offer an improvement over the standard ones is the resulting homogeneity of the trees' leaves in terms of risk level (LGD). Although it is complicated to define homogeneity in the context of a continuous target variable, we choose to adopt a derivative of entropy as its measure, which we calculate on a quantization of the LGD, and adjust to penalize the deviation of observations from the mean of the leaf on top of their scattering. As with the regular entropy, the smaller the value, the more homogeneous the distribution.

\begin{figure}[tbh!]
    \centering
    \includegraphics[width=0.75\columnwidth]{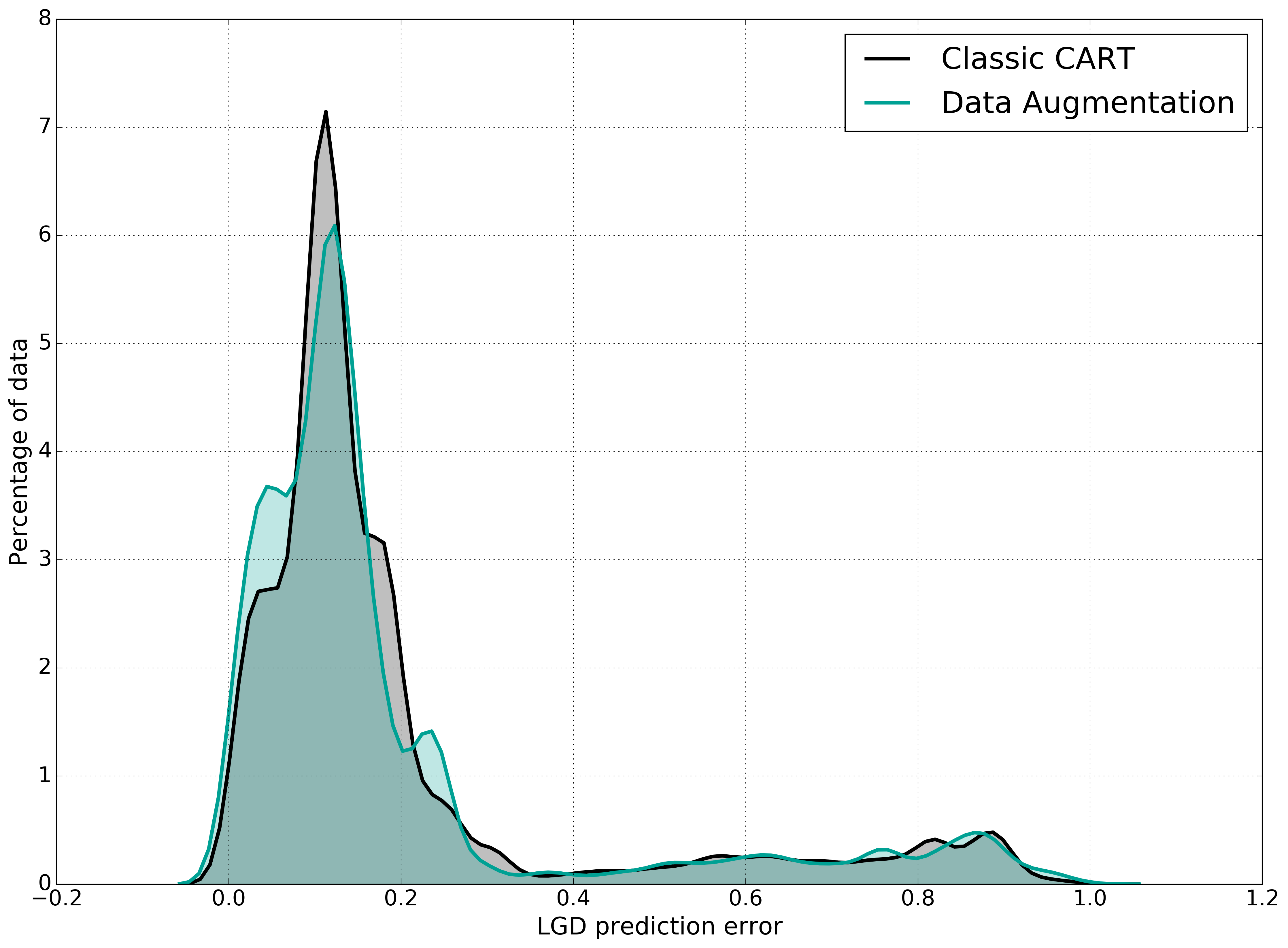}
    \caption{Prediction error distributions for vanilla and student trees trained with a depth of 7.}
    \label{error_distribution}
\end{figure}

As can be observed from Figure \ref{error_distribution} and appendix \ref{results_LGD}, the student model produces a higher proportion of close estimates, with a slightly fatter tail --- i.e. a slightly higher proportion of larger errors. MSE, on the other hand, is equivalent for each pair of models.

\newpage

The mean leaf entropy of the standard and student trees is provided in Table \ref{tab:LGD_homogeneity}. For the same depth, the mean entropy of the student model is remarkably smaller than the standard decision tree's one (6.5 points on average across all depths), which indicates that the student tree is better at gathering observations with a similar target value. It is also expected to see the mean entropy decreasing as the depth of the trees increments, since this increase in complexity allows for thinner cuts.\\

\begin{table}[!htbp]
\caption{Mean leaf homogeneity measure based on an entropy derivative for standard and student models.}
\label{tab:LGD_homogeneity}
\vskip 0.02in
\begin{center}
\begin{tabular}{ccc}
\toprule
& \multicolumn{2}{c}{\textit{\textbf{Mean entropy}}} \\ \cmidrule{2-3}
Depth & Standard  & Student \\ \midrule
4     & 0.422     & \textbf{0.356}  \\
5     & 0.420     & \textbf{0.353}  \\
6     & 0.417     & \textbf{0.350}  \\
7     & 0.414     & \textbf{0.348}  \\
8     & 0.410     & \textbf{0.345}  \\
9     & 0.407     & \textbf{0.342}  \\
10    & 0.403     & \textbf{0.339}  \\
11    & 0.398     & \textbf{0.335}  \\
12    & 0.392     & \textbf{0.329}  \\ \bottomrule
\end{tabular}
\end{center}
\vskip -0.02in
\end{table}

These results illustrate that the use of knowledge distillation also yields promising results in real world applications, for both classification and regression tasks.

\section{Conclusion}
In this paper, we put forward a literature review of knowledge distillation techniques with a focus on those applicable to simple models such as decision trees, which we clustered in three broad categories, a comparison between selected variants on open datasets, and two applications of such methods in a retail banking context. We demonstrated the potential benefits of knowledge distillation in the form of improved discriminatory power without the need to change the model type for students --- thus reaching a middle ground between simple and interpretable models and more complex and accurate ones --- but also its limits. As our experiments have shown, the efficiency of a specific method depends on the use case.

As such, more research is needed to characterize datasets on which knowledge distillation will work on one hand, and what constitutes a good teacher on the other hand. Indeed, this paradigm does not always benefit the most from the highest performing models, but from a good fit between teacher, dataset, and student, allowing for additional knowledge to be transferred. Likewise, some of the approaches we explored are not mutually exclusive, and their combination could lead to further improvements. The results we obtained suggest that simple models are sometimes capable of higher performance but are held back by their training algorithm, offering a glimpse at possible improvements.

\clearpage
\onecolumn

\section*{Acknowledgements}
The infrastructure and technical support of BPCE --- and especially the Regulatory AI department --- is gratefully acknowledged. The authors are thankful to their co-workers for their insightful discussions and patient proofreading, and to the Modeling department for granting access to credit risk related resources and datasets.

\printbibliography

\newpage
\section*{Appendices}

\renewcommand{\thefigure}{A.\arabic{figure}}
\setcounter{figure}{0}

\renewcommand{\thetable}{A.\arabic{table}}
\setcounter{table}{0}

\begin{table*}[!htbp]
\caption{Proportions of training, validation, unlabeled (in the case of the \textbf{SGEMM} dataset) and test samples used in experiments.}
\label{train_test_valid}
\vskip 0.1in
\begin{adjustbox}{center}
\begin{tabular}{ccccc}
\toprule
Dataset & Training & Validation & Unlabeled & Test \\ \midrule
MNIST    & 5/7   & 1/7 & \textit{N/A} & 1/7   \\
Connect4 & 8/10  & 1/10 & \textit{N/A} & 1/10  \\
Adult    & 8/10  & 1/10 & \textit{N/A} & 1/10  \\
SGEMM    & 4/250 & \textit{N/A} & 49/50 & 1/250 \\
\bottomrule
\end{tabular}

\end{adjustbox}
\vskip -0.1in
\end{table*}

\begin{table*}[!htbp]
\caption{Mean \textbf{F1 scores} (in \%) of the standard CART and the student models resulting from different knowledge distillation techniques, with a tree depth ranging from 4 to 12, on the \textbf{Adult} dataset. The F1 scores are provided with the standard deviation (in \%) between parenthesis as all results are the mean of 10 different random splits of the dataset into training, validation and test samples.}
\label{results_adult}
\vskip 0.1in
\begin{adjustbox}{center}
\resizebox{1.2\textwidth}{!}{
\begin{tabular}{cccccccccc}
\toprule
& \textit{Standard} & & \multicolumn{5}{c}{\textit{Soft Targets}} & & \textit{Sample Selection} \\ \cmidrule{2-2} \cmidrule{4-8} \cmidrule{10-10}
Tree depth & CART            & & Vanilla              & Probability Shift    & Label Smoothing      & Mixed Labels & Matching Logits & & ProfWeight \\ \midrule
4          & 61.03 (1.3)     & & \textbf{64.28 (1.5)} & 63.92 (1.2)          & 64.23 (0.8)          & 63.59 (1.0)  & 62.49 (2.6)     & & 62.16 (1.7)\\
5          & 62.95 (1.1)     & & 64.99 (1.4)          & \textbf{66.0 (0.6)}  & 65.45 (1.4)          & 65.04 (1.0)  & 62.78 (1.4)     & & 63.9 (1.0) \\
6          & 64.04 (1.1)     & & 65.73 (0.9)          & \textbf{66.47 (0.8)} & 66.43 (1.3)          & 65.39 (1.0)  & 64.61 (1.3)     & & 66.09 (1.1)\\
7          & 63.66 (1.0)     & & 66.15 (0.8)          & \textbf{66.61 (0.9)} & 66.33 (1.0)          & 65.65 (1.2)  & 65.08 (1.1)     & & 65.79 (1.5)\\
8          & 66.12 (1.1)     & & 67.45 (1.0)          & \textbf{68.26 (0.9)} & 67.78 (0.6)          & 67.23 (0.9)  & 66.01 (1.2)     & & 67.07 (0.5)\\
9          & 65.74 (1.2)     & & \textbf{67.76 (1.4)} & 67.74 (1.3)          & 67.56 (1.2)          & 66.65 (1.3)  & 66.99 (1.4)     & & 66.66 (0.7)\\
10         & 65.97 (1.2)     & & 67.74 (0.9)          & \textbf{68.24 (0.9)} & 68.08 (1.2)          & 66.82 (0.9)  & 67.03 (1.1)     & & 66.91 (1.3)\\
11         & 67.19 (1.0)     & & 68.75 (0.9)          & \textbf{68.78 (0.9)} & 68.62 (0.7)          & 67.39 (1.3)  & 67.82 (1.3)     & & 68.06 (0.9)\\
12         & 66.74 (1.5)     & & 68.52 (1.2)          & 68.51 (1.1)          & \textbf{68.78 (0.8)} & 67.4 (1.4)   & 67.95 (1.3)     & & 67.64 (0.6)\\ \bottomrule
\end{tabular}
}
\end{adjustbox}\textbf{}
\vskip -0.1in
\end{table*}

\begin{table*}[!htbp]
\caption{Mean \textbf{accuracies} (in \%) of the standard CART and the student models resulting from different knowledge distillation techniques, with a tree depth ranging from 4 to 12, on the \textbf{Connect-4} dataset. The accuracies are provided with the standard deviation (in \%) between parenthesis as all results are the mean of 10 different random splits of the dataset into training, validation and test samples.}
\label{results_connect4}
\vskip 0.1in
\begin{adjustbox}{center}
\resizebox{1.2\textwidth}{!}{
\begin{tabular}{cccccccccc}
\toprule
& \textit{Standard} & & \multicolumn{5}{c}{\textit{Soft Targets}} & & \textit{Sample Selection} \\ \cmidrule{2-2} \cmidrule{4-8} \cmidrule{10-10}
Tree depth & CART            & & Vanilla     & Probability Shift & Label Smoothing & Mixed Labels & Matching Logits & & ProfWeight \\ \midrule
4          & 67.11 (1.6)     & & 66.86 (0.3) & 67.41 (0.2)       & 66.86 (0.3)     & 69.17 (0.5)  & 66.27 (0.5)     & & \textbf{69.39 (0.5)}\\
5          & 68.59 (0.8)     & & 66.99 (1.5) & 67.65 (0.4)       & 67.03 (1.4)     & 70.36 (0.5)  & 66.88 (1.0)     & & \textbf{70.73 (0.4)}\\
6          & 69.94 (1.0)     & & 69.36 (0.4) & 69.05 (0.2)       & 69.24 (0.7)     & 71.55 (0.2)  & 68.96 (0.9)     & & \textbf{71.86 (0.3)}\\
7          & 69.48 (0.6)     & & 69.87 (0.5) & 69.95 (0.6)       & 70.0 (0.4)      & 72.12 (0.3)  & 69.97 (0.5)     & & \textbf{72.5 (0.4)}\\
8          & 70.21 (1.3)     & & 70.06 (0.8) & 70.85 (0.5)       & 70.33 (0.4)     & 72.59 (0.7)  & 70.23 (0.7)     & & \textbf{73.23 (0.4)}\\
9          & 70.79 (0.5)     & & 70.97 (0.6) & 71.5 (0.5)        & 70.94 (0.6)     & 73.24 (0.3)  & 70.93 (0.4)     & & \textbf{73.56 (0.3)}\\
10         & 70.9 (0.6)      & & 71.47 (0.4) & 72.06 (0.6)       & 71.32 (0.5)     & 73.77 (0.3)  & 71.35 (0.6)     & & \textbf{74.06 (0.3)}\\
11         & 71.49 (0.6)     & & 71.68 (0.6) & 72.7 (0.3)        & 71.46 (1.1)     & 74.11 (0.3)  & 71.93 (0.7)     & & \textbf{74.55 (0.3)}\\
12         & 71.4 (1.0)      & & 72.27 (0.4) & 73.27 (0.5)       & 72.21 (0.5)     & 74.42 (0.3)  & 72.34 (0.4)     & & \textbf{74.66 (0.4)}\\ \bottomrule
\end{tabular}
}
\end{adjustbox}
\vskip -0.1in
\end{table*}

\begin{table*}[!htbp]
\caption{Mean \textbf{accuracies} (in \%) of the standard CART and the student models resulting from different knowledge distillation techniques, with a tree depth ranging from 4 to 12, on the \textbf{MNIST} dataset. The accuracies are provided with the standard deviation (in \%) between parenthesis as all results are the mean of 10 different random splits of the dataset into training, validation and test samples.}
\label{results_mnist}
\vskip 0.1in
\begin{adjustbox}{center}
\resizebox{1.2\textwidth}{!}{
\begin{tabular}{cccccccccc}
\toprule
& \textit{Standard} & & \multicolumn{5}{c}{\textit{Soft Targets}} & & \textit{Sample Selection} \\ \cmidrule{2-2} \cmidrule{4-8} \cmidrule{10-10}
Tree depth & CART                & & Vanilla              & Probability Shift & Label Smoothing      & Mixed Labels         & Matching Logits     & & ProfWeight \\ \midrule
4          & 62.96 (0.0)         & & \textbf{62.98 (0.0)} & 62.95 (0.0)       & 58.82 (4.1)          & 62.96 (0.0)          & 60.12 (0.0)         & & \textbf{62.98 (0.0)}\\
5          & 65.59 (0.0)         & & 65.53 (0.0)          & 65.51 (0.0)       & \textbf{67.64 (0.1)} & 65.59 (0.0)          & 63.91 (0.0)         & & 65.95 (0.0)\\
6          & 72.56 (0.0)         & & 72.26 (0.0)          & 71.95 (0.0)       & 72.17 (0.2)          & 72.56 (0.0)          & \bf{73.15 (0.0)}    & & 72.28 (0.0)\\
7          & 77.25 (0.0)         & & 77.21 (0.0)          & 76.97 (0.0)       & \textbf{77.79 (0.8)} & 77.17 (0.0)          & 76.95 (0.0)         & & 76.75 (0.0)\\
8          & 79.67 (0.0)         & & 79.82 (0.0)          & 79.89 (0.0)       & 80.84 (1.1)          & 79.82 (0.0)          & \textbf{80.88 (0.0)} & & 79.73 (0.0)\\
9          & 82.88 (0.0)         & & 83.34 (0.0)          & 83.4 (0.0)        & 82.98 (0.1)          & \textbf{83.43 (0.0)} & 82.79 (0.0)         & & 82.77 (0.0)\\
10         & 84.68 (0.0) & & 84.61 (0.0)          & 84.48 (0.0)               & 83.77 (0.6)          & 84.61 (0.0)          & 84.25 (0.0)         & & \bf{85.0 (0.0)} \\
11         & 85.66 (0.0)         & & 86.12 (0.0)          & 86.17 (0.0)       & 84.8 (0.6)           & \textbf{86.18 (0.0)} & 84.45 (0.0)         & & 85.07 (0.0)\\
12         & 85.82 (0.0)         & & \textbf{86.39 (0.0)} & 86.15 (0.0)       & 84.93 (0.5)          & 86.26 (0.0)          & 85.37 (0.0)         & & 86.24 (0.0)\\ \bottomrule
\end{tabular}
}
\end{adjustbox}
\vskip -0.1in
\end{table*}

\begin{table}[!htbp]
\caption{Mean \textbf{MSE} (x$10^{-2}$) of the standard CART and the student models resulting from the Data Augmentation technique with a tree depth ranging from 4 to 10, on the \textbf{SGEMM} dataset. The MSE are provided with the standard deviation (in \%) between parenthesis as all results are the mean of 10 different random splits of the dataset into training and test samples.}
\label{tab:SGEMM_results}
\vskip 0.02in
\begin{center}
\begin{tabular}{ccc}
\toprule
Tree depth & Standard               & Student              \\ \midrule
4          & \textbf{48.63 (5.27)} & 49.77 (5.11)          \\
5          & \textbf{28.70 (4.04)} & 29.10 (2.74)          \\
6          & 23.03 (4.06)          & \textbf{17.42 (1.93)} \\
7          & 22.88 (4.03)          & \textbf{10.13 (1.71)} \\
8          & 22.87 (4.03)          & \textbf{7.14 (1.45)}  \\
9          & 22.87 (4.03)          & \textbf{7.05 (1.42)}  \\
10         & 22.87 (4.03)          & \textbf{7.04 (1.42)}  \\ \bottomrule
\end{tabular}
\end{center}
\vskip -0.02in
\end{table}

\begin{table*}[!htbp]
\caption{Weighted recall and hybrid F1 score of the standard CART, undersampling, oversampling, and the student models stemming from the application of different knowledge distillation techniques, with a tree depth ranging from 4 to 10, for the detection of fraudulent transactions.}
\label{results_fraud}
\vskip 0.1in
\begin{adjustbox}{center}
\resizebox{1.2\textwidth}{!}{
\begin{tabular}{cccccccccccc}
\toprule
 & \multicolumn{5}{c}{\textit{Hybrid F1 Score}} & & \multicolumn{5}{c}{\textit{Weighted Recall}} \\
\cmidrule{2-6} \cmidrule{8-12}
Tree depth & Standard & Undersampling & Oversampling & Soft Targets & ProfWeight & & Standard & Undersampling & Oversampling & Soft Targets & ProfWeight\\ \midrule
4          & 0.301 & 0.387 & 0.398 & \textbf{0.404} & 0.396 & & 0.206 & \textbf{0.331} & 0.315          & 0.329          & 0.323          \\
5          & 0.291 & 0.383 & 0.382 & \textbf{0.405} & 0.396 & & 0.202 & 0.317          & \textbf{0.343} & 0.329          & 0.323          \\
6          & 0.298 & 0.341 & 0.382 & \textbf{0.407} & 0.406 & & 0.210 & 0.261          & 0.316          & 0.329          & \textbf{0.337} \\
7          & 0.260 & 0.353 & 0.337 & \textbf{0.407} & 0.361 & & 0.183 & 0.312          & 0.273          & \textbf{0.329} & 0.278          \\
8          & 0.264 & 0.327 & 0.308 & \textbf{0.405} & 0.352 & & 0.191 & 0.299          & 0.254          & \textbf{0.329} & 0.279          \\
9          & 0.250 & 0.297 & 0.305 & \textbf{0.408} & 0.366 & & 0.188 & 0.309          & 0.260          & \textbf{0.335} & 0.297          \\
10         & 0.251 & 0.239 & 0.263 & \textbf{0.408} & 0.357 & & 0.199 & 0.269          & 0.247          & \textbf{0.335} & 0.282          \\ \bottomrule
\end{tabular}
}
\end{adjustbox}
\vskip -0.1in
\end{table*}

\begin{table*}[!htbp]
\caption{Proportion of observations exhibiting a prediction error respectively under 5\%, 10\% and 20\%. The prediction error is the difference between the ground truth LGD and the predicted one. The table compares the results of the standard CART and those of the student one.}
\label{results_LGD}
\vskip 0.1in
\begin{adjustbox}{center}
\begin{tabular}{ccccccccc}
\toprule
& \multicolumn{2}{c}{\textit{Pred. Error < 0.05}} & & \multicolumn{2}{c}{\textit{Pred. Error < 0.1}} & & \multicolumn{2}{c}{\textit{Pred. Error < 0.2}} \\
\cmidrule{2-3} \cmidrule{5-6} \cmidrule{8-9}
Tree depth & Standard            & Student                   & & Standard               & Student               & & Standard                & Student                \\ \midrule
4          & 25.71               & \textbf{31.76}            & & 37.17                  & \textbf{57.08}        & & \textbf{90.41}         & 88.31                  \\
5          & 26.49               & \textbf{31.93}            & & 38.67                  & \textbf{54.33}        & & \textbf{96.15}         & 94.51                  \\
6          & 27.61               & \textbf{31.83}            & & 39.17                  & \textbf{47.63}        & & \textbf{93.44}         & 90.67                  \\
7          & 27.55               & \textbf{34.66}            & & 39.15                  & \textbf{50.07}        & & \textbf{93.56}         & 89.20                  \\
8          & 28.63               & \textbf{34.57}            & & 49.38                  & \textbf{53.46}        & & \textbf{92.19}         & 91.89                  \\
9          & 29.42               & \textbf{34.67}            & & 49.28                  & \textbf{53.65}        & & 90.77                  & \textbf{91.53}         \\
10         & 30.08               & \textbf{34.74}            & & \textbf{56.29}         & 52.90                 & & \textbf{90.32}         & 88.72                  \\
11         & 30.71               & \textbf{35.71}            & & 54.16                  & \textbf{57.94}        & & \textbf{89.42}         & 89.11                  \\
12         & 31.88               & \textbf{36.88}            & & 57.44                  & \textbf{60.91}        & & \textbf{89.75}         & 89.64                  \\ \bottomrule
\end{tabular}
\end{adjustbox}
\vskip -0.1in
\end{table*}

\begin{figure}[tbh!]
    \centering
    \includegraphics[width=\columnwidth]{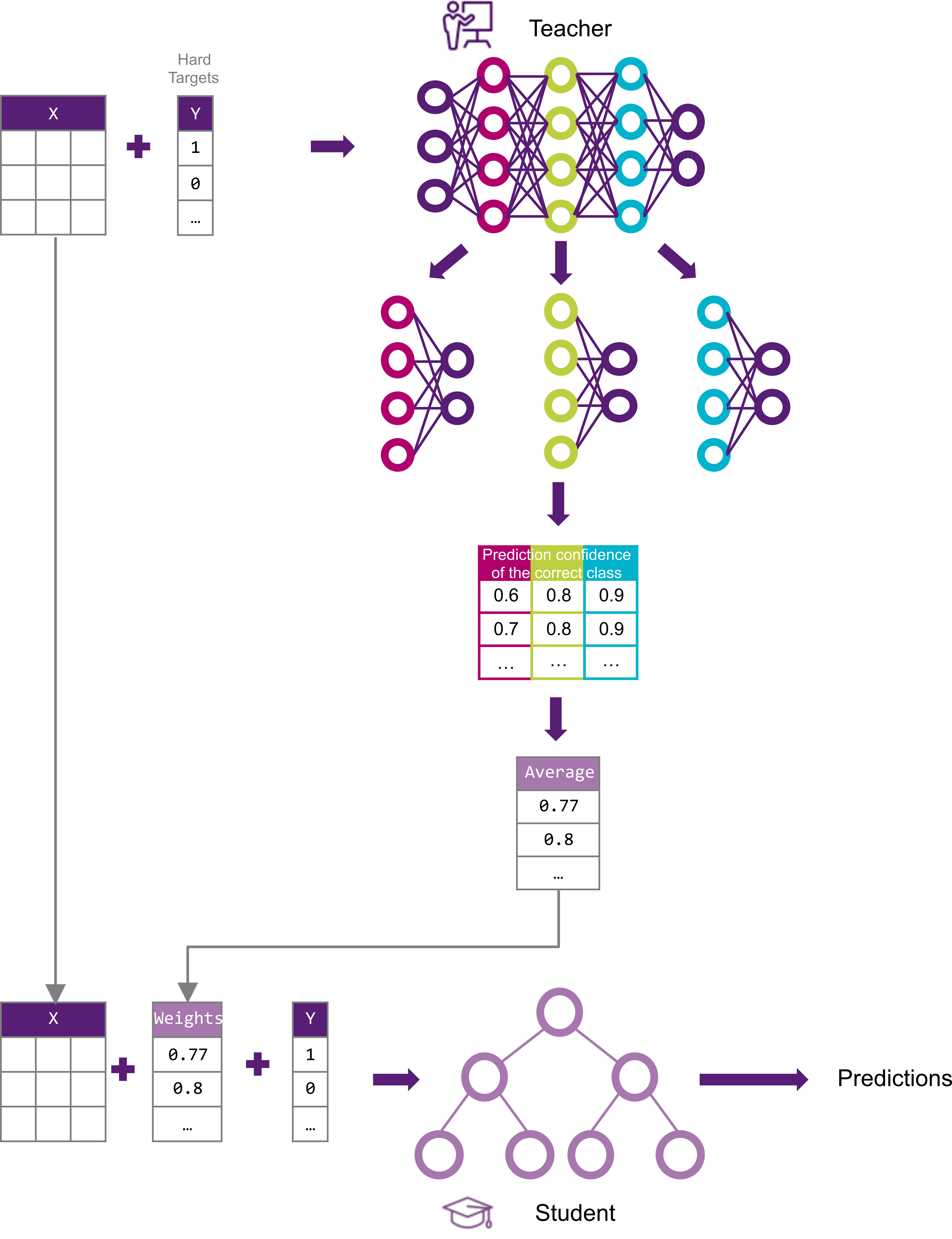}
    \caption{Explicative schema of ProfWeight}
    \label{sample_selection_schema}
\end{figure}

\end{document}